# Generalized Instrumental Variables

**Carlos Brito and Judea Pearl**
Cognitive Systems Laboratory
Computer Science Department
University of California, Los Angeles, CA 90024
*fisch@cs.ucla.edu judea@cs.ucla.edu*

## Abstract

This paper concerns the assessment of direct causal effects from a combination of: (i) non-experimental data, and (ii) qualitative domain knowledge. Domain knowledge is encoded in the form of a directed acyclic graph (DAG), in which all interactions are assumed linear, and some variables are presumed to be unobserved. We provide a generalization of the well-known method of Instrumental Variables, which allows its application to models with few conditional independeces.

## 1 Introduction

This paper explores the feasibility of inferring linear cause-effect relationships from various combinations of data and theoretical assumptions. The assumptions are represented in the form of an acyclic causal diagram which contains both arrows and bi-directed arcs [9, 10]. The arrows represent the potential existence of direct causal relationships between the corresponding variables, and the bi-directed arcs represent spurious correlations due to unmeasured common causes. All interactions among variables are assumed to be linear. Our task is to decide whether the assumptions represented in the diagram are sufficient for assessing the strength of causal effects from non-experimental data, and, if sufficiency is proven, to express the target causal effect in terms of estimable quantities.

This decision problem has been tackled in the past half century, primarily by econometricians and social scientists, under the rubric "The Identification Problem" [6] – it is still unsolved. Certain restricted classes of models are nevertheless known to be identifiable, and these are often assumed by social scientists as a matter of convenience or convention [5]. A hierarchy of three such classes is given in [7]: (1) no bidirected arcs, (2) bidirected arcs restricted to root variables,

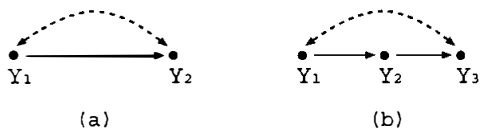

Figure 1: (*a*) a "bow-pattern", and (*b*) a bow-free model

and (3) bidirected arcs restricted to variables that are not connected through directed paths.

Recently [4], we have shown that the identification of the entire model is ensured if variables standing in direct causal relationship (i.e., variables connected by arrows in the diagram) do not have correlated errors; no restrictions need to be imposed on errors associated with indirect causes. This class of models was called "bow-free", since their associated causal diagrams are free of any "bow pattern" [10] (see Figure 1).

Most existing conditions for Identification in general models are based on the concept of Instrumental Variables (IV) [11], [2]. IV methods take advantage of conditional independence relations implied by the model to prove the Identification of specific causal-effects. When the model is not rich in conditional independences, these methods are not much informative. In [3], we proposed a new graphical criterion for Identification which does not make direct use of conditional independence, and thus can be successfully applied to models in which IV methods would fail.

In this paper, we provide an important generalization of the method of Instrumental Variables that makes it less sensitive to the independence relations implied by the model.

## 2 Linear Models and Identification

An equation $Y = \beta X + e$ encodes two distinct assumptions: (1) the possible existence of (direct) causal influence of $X$ on $Y$; and, (2) the absence of causal in-

$$Z = e_1$$
$$W = e_2$$
$$X = aZ + e_3$$
$$Y = bW + cX + e_4$$
$$Cov(e_1, e_2) = \alpha \neq 0$$
$$Cov(e_2, e_3) = \beta \neq 0$$
$$Cov(e_3, e_4) = \gamma \neq 0$$

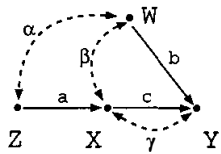

Figure 2: A simple linear model and its causal diagram

fluence on $Y$ of any variable that does not appear on the right-hand side of the equation. The parameter $\beta$ quantifies the (direct) causal effect of $X$ on $Y$. That is, the equation claims that a unit increase in $X$ would result in $\beta$ units increase of $Y$, assuming that everything else remains the same. The variable $e$ is called an "error" or "disturbance"; it represents unobserved background factors that the modeler decides to keep unexplained.

A linear model for a set of random variables $\mathbf{Y} = \{Y_1, \ldots, Y_n\}$ is defined formally by a set of equations of the form

$$Y_j = \sum_i c_{ji} Y_i + e_j \qquad , j = 1, \ldots, n$$

and an error variance/covariance matrix $\Psi$, i.e., $[\Psi_{ij}] = Cov(e_i, e_j)$. The error terms $e_j$ are assumed to have normal distribution with zero mean.

The equations and the pairs of error-terms $(e_i, e_j)$ with non-zero correlation define the structure of the model. The model structure can be represented by a directed graph, called causal diagram, in which the set of nodes is defined by the variables $Y_1, \ldots, Y_n$, and there is a directed edge from $Y_i$ to $Y_j$ if the coefficient of $Y_i$ in the equation for $Y_j$ is different from zero. Additionally, if error-terms $e_i$ and $e_j$ have non-zero correlation, we add a (dashed) bidirected edge between $Y_i$ and $Y_j$. Figure 2 shows a model with the respective causal diagram.

The structural parameters of the model, denoted by $\theta$, are the coefficients $c_{ij}$, and the non-zero entries of the error covariance matrix $\Psi$. In this work, we consider only recursive models, that is, $c_{ji} = 0$ for $i \geq j$.

Fixing the model structure and assigning values to the parameters $\theta$, the model determines a unique covariance matrix $\Sigma$ over the observed variables $\{Y_1, \ldots, Y_n\}$, given by (see [1], page 85)

$$\Sigma(\theta) = (I - C)^{-1} \Psi \left[ (I - C)^{-1} \right]^T \qquad (1)$$

where $C$ is the matrix of coefficients $c_{ji}$.

Conversely, in the Identification problem, after fixing the structure of the model, one attempts to solve for $\theta$ in terms of the observed covariance $\Sigma$. This is not always possible. In some cases, no parametrization of the model could be compatible with a given $\Sigma$. In other cases, the structure of the model may permit several distinct solutions for the parameters. In these cases, the model is called *nonidentified*.

Sometimes, although the model is nonidentifiable, some parameters may be uniquely determined by the given assumptions and data. Whenever this is the case, the specific parameters are *identified*.

Finally, since the conditions we seek involve the structure of the model alone, and do not depend on the numerical values of parameters $\theta$, we insist only on having identification almost everywhere, allowing few pathological exceptions. The concept of identification almost everywhere is formalized in section 6.

## 3   Graph Background

**Definition 1** *A path in a graph is a sequence of edges (directed or bidirected) such that each edge starts in the node ending the preceding edge. A directed path is a path composed only by directed edges, all oriented in the same direction. Node $X$ is a* descendent *of node $Y$ if there is a directed path from $Y$ to $X$. Node $Z$ is a collider in a path $p$ if there is a pair of consecutive edges in $p$ such that both edges are oriented toward $Z$ (e.g.,$\ldots \to Z \leftarrow \ldots$).*

Let $p$ be a path between $X$ and $Y$, and let $Z$ be an intermediate variable in $p$. We denote by $p[X \sim Z]$ the subpath of $p$ consisting of the edges between $X$ and $Z$.

**Definition 2** (*d-separation*)
*A set of nodes $\mathbf{Z}$ d-separates $X$ from $Y$ in a graph, if $Z$ blocks every path between $X$ and $Y$. A path $p$ is blocked by a set $\mathbf{Z}$ (possibly empty) if one of the following holds:*

(i) *$p$ contains at least one non-collider that is in $\mathbf{Z}$;*

(ii) *$p$ contains at least one collider that is outside $\mathbf{Z}$ and has no descendant in $\mathbf{Z}$.*

## 4   Instrumental Variable Methods

The traditional definition qualifies a variable $Z$ as instrumental, relative to a cause $X$ and effect $Y$ if [10]:

1. $Z$ is independent of all error terms that have an influence on $Y$ which is not mediated by $X$;

2. $Z$ is not independent of $X$.

The intuition behind this definition is that all correlation between $Z$ and $Y$ must be intermediated by $X$.

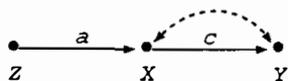

Figure 3: Typical Instrumental Variable

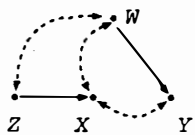

Figure 4: Conditional IV Examples

If we can find $Z$ with these properties, then the causal effect of $X$ on $Y$, denoted by $c$, is identified and given by $c = \sigma_{ZY}/\sigma_{ZX}$.

Figure 3 shows a typical example of an instrumental variable. It is easy to verify that variable $Z$ satisfy properties (1) and (2) in this model.

A generalization of the IV method is offered through the use of conditional IV's. A conditional IV is a variable $Z$ that may not have properties (1) and (2), but there is a conditioning set $\mathbf{W}$ that makes it happen. When such pair $(Z, \mathbf{W})$ is found, the causal effect of $X$ on $Y$ is identified and given by $c = \sigma_{ZY.\mathbf{W}}/\sigma_{ZX.\mathbf{W}}$.

[11] provides the following equivalent graphical criterion for conditional IV's, based on the concept of d-separation:

1. $\mathbf{W}$ contains only non-descendents of $Y$;

2. $\mathbf{W}$ d-separates $Z$ from $Y$ in the subgraph $G_c$ obtained by removing edge $X \to Y$ from $G$;

3. $\mathbf{W}$ does not d-separate $Z$ from $X$ in $G_c$.

As an example of the application of this criterion, Figure 4 show the graph obtained by removing edge $X \to Y$ from the model of Figure 2. After conditioning on variable $W$, $Z$ becomes d-separated from $Y$ but not from $X$. Thus, parameter $c$ is identified.

## 5 Instrumental Sets

Although very useful, the method of conditional IV's has some limitations. As an example, Figure (5a) shows a simple model in which the method cannot be applied. In this model, variables $Z_1$ and $Z_2$ do not qualify as IV's with respect to either $c_1$ or $c_2$. Also, there is no conditioning set which makes it happen. Therefore, the conditional IV method fails, despite the fact that the model is completely identified.

Following the ideas stated in the graphical criterion for conditional IV's, we show in Figure (5b) the graph

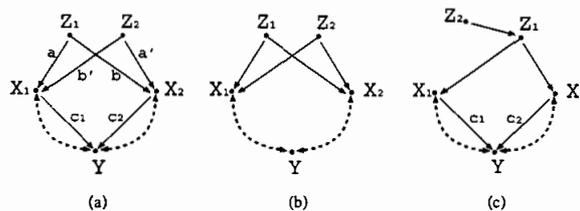

Figure 5: Simultaneous use of two IVs

obtained by removing edges $X_1 \to Y$ and $X_2 \to Y$ from the model. Note that in this graph, $Z_1$ and $Z_2$ satisfy the graphical conditions for a conditional IV. Intuitively, if we could use both $Z1$ and $Z_2$ together as instrumental variables, we would be able to identify parameters $c_1$ and $c_2$. This motivates the following informal definition:

A set of variables $\mathbf{Z} = \{Z_1, \ldots, Z_k\}$ is called an *instrumental set* relative to a set of causes $\mathbf{X} = \{X_1, \ldots, X_n\}$ and an effect $Y$ if:

1. Each $Z_i \in \mathbf{Z}$ is independent of all error terms that have an influence on $Y$ which is not mediated by some $X_j \in \mathbf{X}$;

2. Each $Z_i \in \mathbf{Z}$ is not independent of the respective $X_i \in \mathbf{X}$, for appropriate enumerations of $\mathbf{Z}$ and $\mathbf{X}$;

3. The set $\mathbf{Z}$ is not redundant with respect to $Y$. That is, for any $Z_i \in \mathbf{Z}$ we cannot explain the correlation between $Z_i$ and $Y$ by correlations between $Z_i$ and $\mathbf{Z} - \{Z_i\}$, and correlations between $\mathbf{Z} - \{Z_i\}$ and $Y$.

Properties 1 and 2 above are similar to the ones in the definition of Instrumental Variables, and property 3 is required when using more than one instrument. To see why we need the extra condition, let us consider the model in Figure (5c). In this example, the correlation between $Z_2$ and $Y$ is given by the product of the correlation between $Z_2$ and $Z_1$ and the correlation between $Z_1$ and $Y$. That is, $Z_2$ does not give additional information once we already have $Z_1$. In fact, using $Z_1$ and $Z_2$ as instruments we cannot obtain the identification of the causal effects of $X_1$ and $X_2$ on $Y$.

Now, we give a precise definition of instrumental sets using graphical conditions. Fix a variable $Y$ and let $\mathbf{X} = \{X_1, \ldots, X_k\}$ be a set of direct causes of $Y$.

**Definition 3** *The set* $\mathbf{Z} = \{Z_1, \ldots, Z_n\}$ *is said to be an Instrumental Set relative to* $\mathbf{X}$ *and* $Y$ *if we can find triples* $(Z_1, \mathbf{W}_1, p_1), \ldots, (Z_n, \mathbf{W}_n, p_n)$, *such that:*

*(i) For* $i = 1, \ldots, n$, $Z_i$ *and the elements of* $\mathbf{W}_i$

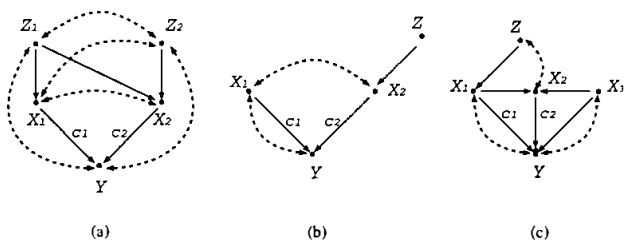

Figure 6: More examples of Instrumental Sets

*are non-descendents of $Y$; and $p_i$ is an unblocked path between $Z_i$ and $Y$ including edge $X_i \rightarrow Y$.*

*(ii) Let $\overline{G}$ be the causal graph obtained from $G$ by deleting edges $X_1 \rightarrow Y, \ldots, X_n \rightarrow Y$. Then, $\mathbf{W}_i$ d-separates $Z_i$ from $Y$ in $\overline{G}$; but $\mathbf{W}_i$ does not block path $p_i$;*

*(iii) For $1 \le i < j \le n$, variable $Z_j$ does not appear in path $p_i$; and, if paths $p_i$ and $p_j$ have a common variable $V$, then both $p_i[V \sim Y]$ and $p_j[Z_j \sim V]$ point to $V$.*

Next, we state the main result of this paper.

**Theorem 1** *If $\mathbf{Z} = \{Z_1, \ldots, Z_n\}$ is an instrumental set relative to causes $\mathbf{X} = \{X_1, \ldots, X_n\}$ and effect $Y$, then the parameters of edges $X_1 \rightarrow Y, \ldots, X_n \rightarrow Y$ are identified almost everywhere, and can be computed by solving a system of linear equations.*

Figure 6 shows more examples in which the method of conditional IV's fails and our new criterion is able to prove the identification of parameters $c_i$'s. In particular, model $(a)$ is a bow-free model, and thus is completely identifiable. Model $(b)$ illustrates an interesting case in which variable $X_2$ is used as the instrument for $X_1 \rightarrow Y$, while $Z$ is the instrument for $X_2 \rightarrow Y$. Finally, in model $(c)$ we have an example in which the parameter of edge $X_3 \rightarrow Y$ is nonidentifiable, and still the method can prove the identification of $c_1$ and $c_2$.

The remaining of the paper is dedicated to the proof of Theorem 1.

## 6    Preliminary Results

### 6.1    Identification Almost Everywhere

Let $h$ denote the total number of parameters in model $G$. Then, each vector $\theta \in \Re^h$ defines a parametrization of the model. For each parametrization $\theta$, model $G$ generates a unique covariance matrix $\Sigma(\theta)$. Let $\theta(\lambda_1, \ldots, \lambda_n)$ denote the vector of values assigned by $\theta$ to parameters $\lambda_1, \ldots, \lambda_n$.

Parameters $\lambda_1, \ldots, \lambda_n$, are identified almost everywhere if $\Sigma(\theta) = \Sigma(\theta')$ implies $\theta(\lambda_1, \ldots, \lambda_n) = \theta'(\lambda_1, \ldots, \lambda_n)$, except when $\theta$ resides on a set of Lebesgue measure zero.

### 6.2    Wright's Method of Path Coefficients

Here, we describe an important result introduced by Sewall Wright [12], which is extensively explored in the proof.

Given variables $X$ and $Y$ in a recursive linear model, the correlation coefficient of $X$ and $Y$, denoted $\rho_{XY}$, can be expressed as a polynomial on the parameters of the model. More precisely,

$$\rho_{Z,Y} = \sum_{\text{paths } p_l} T(p_l) \qquad (2)$$

where term $T(p_l)$ represents the multiplication of the parameters of edges along path $p_l$, and the summation ranges over all unblocked paths between $X$ and $Y$. For this equality to hold, the variables in the model must be standardized (variance equal to 1) and have zero mean. However, if this is not the case, a simple transformation can put the model in this form [13]. We refer to Eq.(2) as Wright's Equation for $X$ and $Y$.

Wright's method of path coefficients [12] consists in forming Eq.(2) for each pair of variables in the model, and solving for the parameters in terms of the correlations among the variables. Whenever there is a unique solution for a parameter $\lambda$, this parameter is identified.

We can use this method to study the identification of the parameters in the model of Figure 5. From the equations for $\rho_{Y_1,Y_5}$ and $\rho_{Y_2,Y_5}$ we can see that parameters $c_1$ and $c_2$ are identified if and only if

$$Det \begin{bmatrix} a & a' \\ b & b' \end{bmatrix} \ne 0$$

### 6.3    Partial Correlation Lemma

Next lemma provides a convenient expression for the partial correlation coefficient of $Y_1$ and $Y_2$, given $Y_3, \ldots, Y_n$, denoted $\rho_{12.3\ldots n}$. The proof of the lemma is given in the appendix.

**Lemma 1** *The partial correlation $\rho_{12.3\ldots n}$ can be expressed as the ratio:*

$$\rho_{12.3\ldots n} = \frac{\phi(1, 2, \ldots, n)}{\psi(1, 3, \ldots, n) \cdot \psi(2, 3, \ldots, n)} \qquad (3)$$

*where $\phi$ and $\psi$ are functions of the correlations among $Y_1, Y_2, \ldots, Y_n$, satisfying the following conditions:*

*(i)  $\phi(1, 2, \ldots, n) = \phi(2, 1, \ldots, n)$.*

(ii) $\phi(1, 2, \ldots, n)$ is linear on the correlations $\rho_{12}, \rho_{32}, \ldots, \rho_{n2}$, with no constant term.

(iii) The coefficients of $\rho_{12}, \rho_{32}, \ldots, \rho_{n2}$, in $\phi(1, 2, \ldots, n)$ are polynomials on the correlations among the variables $Y_1, Y_3, \ldots, Y_n$. Moreover, the coefficient of $\rho_{12}$ has the constant term equal to 1, and the coefficients of $\rho_{32}, \ldots, \rho_{n2}$, are linear on the correlations $\rho_{13}, \rho_{14}, \ldots, \rho_{1n}$, with no constant term.

(iv) $(\psi(i_1, \ldots, i_{n-1}))^2$, is a polynomial on the correlations among the variables $Y_{i_1}, \ldots, Y_{i_{n-1}}$, with constant term equal to 1.

## 6.4 Path Lemmas

The following lemmas explore some consequences of the conditions in the definition of Instrumental Sets.

**Lemma 2** W.l.o.g., we may assume that, for $1 \leq i < j \leq n$, paths $p_i$ and $p_j$ do not have any common variable other than (possibly) $Z_i$.

**Proof:** Assume that paths $p_i$ and $p_j$ have some variables in common, different from $Z_i$. Let $V$ be the closest variable to $X_i$ in path $p_i$ which also belongs to path $p_j$.

We show that after replacing triple $(Z_i, \mathbf{W}_i, p_i)$ by triple $(V, \mathbf{W}_i, p_i[V \sim Y])$, conditions $(i) - (iii)$ still hold.

It follows from condition $(iii)$ that subpath $p_i[V \sim Y]$ must point to $V$. Since $p_i$ is unblocked, subpath $p_i[Z_i \sim V]$ must be a directed path from $V$ to $Z_i$.

Now, variable $V$ cannot be a descendent of $Y$, because $p_i[Z_i \sim V]$ is a directed path from $V$ to $Z_i$, and $Z_i$ is a non-descendent of $Y$. Thus, condition $(i)$ still holds.

Consider the causal graph $\overline{G}$. Assume that there exists a path $p$ between $V$ and $Y$ witnessing that $\mathbf{W}_i$ does not d-separate $V$ from $Y$ in $\overline{G}$. Since $p_i[Z_i \sim V]$ is a directed path from $V$ to $Z_i$, we can always find another path witnessing that $\mathbf{W}_i$ does not d-separate $Z_i$ from $Y$ in $\overline{G}$ (for example, if $p$ and $p_i[Z_i \sim V]$ do not have any variable in common other than $V$, then we can just take their concatenation). But this is a contradiction, and thus it is easy to see that condition $(ii)$ still holds.

Condition $(iii)$ follows from the fact that $p_i[V \sim Y]$ and $p_j[Z_j \sim V]$ point to $V$.                                  □

In the following, we assume that the conditions of lemma 2 hold.

**Lemma 3** For all $1 \leq i \leq n$, there exists no unblocked path between $Z_i$ and $Y$, different from $p_i$, which includes edge $X_i \rightarrow Y$ and is composed only by edges from $p_1, \ldots, p_i$.

**Proof:** Let $p$ be an unblocked path between $Z_i$ and $Y$, different from $p_i$, and assume that $p$ is composed only by edges from $p_1, \ldots, p_i$.

According to condition $(iii)$, if $Z_i$ appears in some path $p_j$, with $j \neq i$, then it must be that $j > i$. Thus, $p$ must start with some edges of $p_i$.

Since $p$ is different from $p_i$, it must contain at least one edge from $p_1, \ldots, p_{i-1}$. Let $(V_1, V_2)$ denote the first edge in $p$ which does not belong to $p_i$.

From lemma 2, it follows that variable $V_1$ must be a $Z_k$ for some $k < i$, and by condition $(iii)$, subpath $p[Z_i \sim V_1]$ and edge $(V_1, V_2)$ must point to $V_1$. But this implies that $p$ is blocked by $V_1$, which contradicts our assumptions.                                  □

The proofs for the next two lemmas are very similar to the previous one, and so are omitted.

**Lemma 4** For all $1 \leq i \leq n$, there is no unblocked path between $Z_i$ and some $W_{i_j}$ composed only by edges from $p_1, \ldots, p_i$.

**Lemma 5** For all $1 \leq i \leq n$, there is no unblocked path between $Z_i$ and $Y$ including edge $X_j \rightarrow Y$, with $j < i$, composed only by edges from $p_1, \ldots, p_i$.

# 7  Proof of Theorem 1

## 7.1  Notation

Fix a variable $Y$ in the model. Let $\mathbf{X} = \{X_1, \ldots, X_k\}$ be the set of all non-descendents of $Y$ which are connected to $Y$ by an edge (directed, bidirected, or both). Define the following set of edges with an arrowhead at $Y$:

$$Inc(Y) = \{(X_i, Y) : X_i \in \mathbf{X}\}$$

Note that for some $X_i \in \mathbf{X}$ there may be more than one edge between $X_i$ and $Y$ (one directed and one bidirected). Thus, $|Inc(Y)| \geq |\mathbf{X}|$. Let $\lambda_1, \ldots, \lambda_m$, $m \geq k$, denote the parameters of the edges in $Inc(Y)$.

It follows that edges $X_1 \rightarrow Y, \ldots, X_n \rightarrow Y$, belong to $Inc(Y)$, because $X_1, \ldots, X_n$, are clearly non-descendents of $Y$. W.l.o.g., let $\lambda_i$ be the parameter of edge $X_i \rightarrow Y$, $1 \leq i \leq n$, and let $\lambda_{n+1}, \ldots, \lambda_m$ be the parameters of the remaining edges in $Inc(Y)$.

Let $Z$ be any non-descendent of $Y$. Wright's equation for the pair $(Z, Y)$, is given by

$$\rho_{Z,Y} = \sum_{\text{paths } p_l} T(p_l) \qquad (4)$$

where each term $T(p_l)$ corresponds to an unblocked path between $Z$ and $Y$. Next lemma proves a property of such paths.

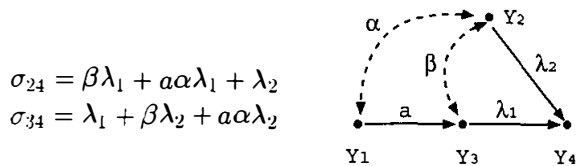

$$\sigma_{24} = \beta\lambda_1 + a\alpha\lambda_1 + \lambda_2$$
$$\sigma_{34} = \lambda_1 + \beta\lambda_2 + a\alpha\lambda_2$$

Figure 7: Wright's equations

**Lemma 6** *Let $Y$ be a variable in a recursive model, and let $Z$ be a non-descendent of $Y$. Then, any unblocked path between $Z$ and $Y$ must include exactly one edge from $Inc(Y)$.*

Lemma 6 allows us to write Eq. (4) as

$$\rho_{Z,Y} = \sum_{j=1}^{m} a_j \cdot \lambda_j \qquad (5)$$

Thus, the correlation between $Z$ and $Y$ can be expressed as a linear function of the parameters $\lambda_1, \ldots, \lambda_m$, with no constant term. Figure 7 shows an example of those equations for a simple model.

### 7.2 Basic Linear Equations

Consider a triple $(Z_i, \mathbf{W}_i, p_i)$, and let $\mathbf{W}_i = \{W_{i_1}, \ldots, W_{i_k}\}$ [1]. From lemma 1, we can express the partial correlation of $Z_i$ and $Y$ given $\mathbf{W}_i$ as:

$$\rho_{Z_i Y \cdot \mathbf{W}_i} = \frac{\phi_i(Z_i, Y, W_{i_1}, \ldots, W_{i_k})}{\psi_i(Z_i, W_{i_1}, \ldots, W_{i_k}) \cdot \psi_i(Y, W_{i_1}, \ldots, W_{i_k})} \qquad (6)$$

where function $\phi_i$ is linear on the correlations $\rho_{Z_i Y}$, $\rho_{W_{i_1} Y}, \ldots, \rho_{W_{i_k} Y}$, and $\psi_i$ is a function of the correlations among the variables given as arguments. We abbreviate $\phi_i(Z_i, Y, W_{i_1}, \ldots, W_{i_k})$ by $\phi_i(Z_i, Y, \mathbf{W}_i)$, and $\psi_i(V, W_{i_1}, \ldots, W_{i_k})$ by $\psi_i(V, \mathbf{W}_i)$.

We have seen that the correlations $\rho_{Z_i Y}, \rho_{W_{i_1} Y}, \ldots, \rho_{W_{i_k} Y}$, can be expressed as linear functions of the parameters $\lambda_1, \ldots, \lambda_m$. Since $\phi_i$ is linear on these correlations, it follows that we can express $\phi_i$ as a linear function of the parameters $\lambda_1, \ldots, \lambda_m$.

Formally, by lemma 1, $\phi_i(Z_i, Y, \mathbf{W}_i)$ can be written as:

$$\phi_i(Z_i, Y, \mathbf{W}_i) = b_{i_0}\rho_{Z_i Y} + b_{i_1}\rho_{W_{i_1} Y} + \ldots + b_{i_k}\rho_{W_{i_k} Y} \qquad (7)$$

Also, for each $V_j \in \{Z_i\} \cup \mathbf{W}_i$ we can write:

$$\rho_{V_j Y} = a_{i_j 1}\lambda_1 + \ldots + a_{i_j m}\lambda_m \qquad (8)$$

[1] To simplify the notation, we assume that $|\mathbf{W}_i| = k$, for $i = 1, \ldots, n$

Replacing each correlation in Eq.(7) by the expression given by Eq. (8), we obtain

$$\phi_i(Z_i, Y, \mathbf{W}_i) = q_{i1}\lambda_1 + \ldots + q_{im}\lambda_m \qquad (9)$$

where the coefficients $q_{il}$'s are given by:

$$q_{il} = \sum_{j=0}^{k} b_{i_j} a_{i_j l} \qquad , l = 1, \ldots, m \qquad (10)$$

**Lemma 7** *The coefficients $q_{i,n+1}, \ldots, q_{im}$ in Eq. (9) are identically zero.*

**Proof:** The fact that $\mathbf{W}_i$ d-separates $Z_i$ from $Y$ in $\overline{G}$, implies that $\rho_{Z_i Y \cdot \mathbf{W}_i} = 0$ in any probability distribution compatible with $\overline{G}$ ([10], pg. 142). Thus, $\phi_i(Z_i, Y, \mathbf{W}_i)$ must vanish when evaluated in $\overline{G}$. But this implies that the coefficient of each of the $\lambda_i$'s in Eq. (9) must be identically zero.

Now, we show that the only difference between evaluations of $\phi_i(Z_i, Y, \mathbf{W}_i)$ on the causal graphs $\overline{G}$ and $G$, consists on the coefficients of parameters $\lambda_1, \ldots, \lambda_n$.

First, observe that coefficients $b_{i_0}, \ldots, b_{i_k}$ are polynomials on the correlations among the variables $Z_i, W_{i_1}, \ldots, W_{i_k}$. Thus, they only depend on the unblocked paths between such variables in the causal graph. However, the insertion of edges $X_1 \rightarrow Y, \ldots, X_n \rightarrow Y$, in $\overline{G}$ does not create any new unblocked path between any pair of $Z_i, W_{i_1}, \ldots, W_{i_k}$ (and obviously does not eliminate any existing one). Hence, the coefficients $b_{i_0}, \ldots, b_{i_k}$ have exactly the same value in the evaluations of $\phi_i(Z_i, Y, \mathbf{W}_i)$ on $\overline{G}$ and $G$.

Now, let $\lambda_l$ be such that $l > n$, and let $V_j \in \{Z_i\} \cup \mathbf{W}_i$. Note that the insertion of edges $X_1 \rightarrow Y, \ldots, X_n \rightarrow Y$, in $\overline{G}$ does not create any new unblocked path between $V_j$ and $Y$ including the edge whose parameter is $\lambda_l$ (and does not eliminate any existing one). Hence, coefficients $a_{i_j l}$, $j = 0, \ldots, k$, have exactly the same value on $\overline{G}$ and $G$.

From the two previous facts, we conclude that, for $l > n$, the coefficient of $\lambda_l$ in the evaluations of $\phi_i(Z_i, Y, \mathbf{W}_i)$ on $\overline{G}$ and $G$ have exactly the same value, namely zero. Next, we argue that $\phi_i(Z_i, Y, \mathbf{W}_i)$ does not vanish when evaluated on $G$.

Finally, let $\lambda_l$ be such that $l \leq n$, and let $V_j \in \{Z_i\} \cup \mathbf{W}_i$. Note that there is no unblocked path between $V_j$ and $Y$ in $\overline{G}$ including edge $X_l \rightarrow Y$, because this edge does not exist in $\overline{G}$. Hence, the coefficient of $\lambda_l$ in the expression for the correlation $\rho_{V_j Y}$ on $\overline{G}$ must be zero.

On the other hand, the coefficient of $\lambda_l$ in the same expression on $G$ is not necessarily zero. In fact, it follows from the conditions in the definition of Instrumental sets that, for $l = i$, the coefficient of $\lambda_i$ contains the term $T(p_i)$.                                    $\square$

From lemma 7, we get that $\phi_i(Z_i, Y, \mathbf{W}_i)$ is a linear function only on the parameters $\lambda_1, \ldots, \lambda_n$.

### 7.3 System of Equations $\Phi$

Rewriting Eq.(6) for each triple $(Z_i, \mathbf{W}_i, p_i)$, we obtain the following system of linear equations on the parameters $\lambda_1, \ldots, \lambda_n$:

$$\Phi = \begin{cases} \phi_1(Z_1, Y, \mathbf{W}_1) = & \rho_{Z_1 Y . \mathbf{W}_1} \\ & \cdot \, \psi_1(Z_1, \mathbf{W}_1) \cdot \psi_1(Y, \mathbf{W}_1) \\ \ldots \\ \phi_n(Z_n, Y, \mathbf{W}_n) = & \rho_{Z_n Y . \mathbf{W}_n} \\ & \cdot \, \psi_n(Z_n, \mathbf{W}_n) \cdot \psi_n(Y, \mathbf{W}_n) \end{cases}$$

where the terms on the right-hand side can be computed from the correlations among the variables $Y, Z_i, W_{i_1}, \ldots, W_{i_k}$, estimated from data.

Our goal is to show that $\Phi$ can be solved uniquely for the $\lambda_i$'s, and so prove the identification of $\lambda_1, \ldots, \lambda_n$. Next lemma proves an important result in this direction. Let $Q$ denote the matrix of coefficients of $\Phi$.

**Lemma 8** $Det(Q)$ is a non-trivial polynomial on the parameters of the model.

**Proof:** From Eq.(10), we get that each entry $q_{il}$ of $Q$ is given by

$$q_{il} = \sum_{j=0}^{k} b_{i_j} \cdot a_{i_j l}$$

where $b_{i_j}$ is the coefficient of $\rho_{W_{i_j} Y}$ (or $\rho_{Z_i Y}$, if $j = 0$), in the linear expression for $\phi_i(Z_i, Y, \mathbf{W}_i)$ in terms of correlations (see Eq.(7)); and $a_{i_j l}$ is the coefficient of $\lambda_l$ in the expression for the correlation $\rho_{W_{i_j} Y}$ in terms of the parameters $\lambda_1, \ldots, \lambda_m$ (see Eq.(8)).

From property $(iii)$ of lemma 1, we get that $b_{i_0}$ has constant term equal to 1. Thus, we can write $b_{i_0} = 1 + \hat{b}_{i_0}$, where $\hat{b}_{i_0}$ represent the remaining terms of $b_{i_0}$.

Also, from condition $(i)$ of Theorem 1, it follows that $a_{i_0 i}$ contains term $T(p_i)$. Thus, we can write $a_{i_0 i} = T(p_i) + \hat{a}_{i_0 i}$, where $\hat{a}_{i_0 i}$ represents all the remaining terms of $a_{i_0 i}$.

Hence, a diagonal entry $q_{ii}$ of $Q$, can be written as

$$q_{ii} = T(p_i)[1 + \hat{b}_{i_0}] \; + \; \hat{a}_{i_0 i} \cdot b_{i_0} \; + \; \sum_{j=1}^{k} b_{i_j} \cdot a_{i_j i} \quad (11)$$

Now, the determinant of $Q$ is defined as the weighted sum, for all permutations $\pi$ of $\langle 1, \ldots, n \rangle$, of the product of the entries selected by $\pi$ (entry $q_{ul}$ is selected by permutation $\pi$ if the $i^{th}$ element of $\pi$ is $l$), where the weights are 1 or $(-1)$, depending on the parity of the permutation. Then, it is easy to see that the term

$$T^\star = \prod_{j=1}^{n} T(p_j)$$

appears in the product of permutation $\pi = \langle 1, \ldots, n \rangle$, which selects all the diagonal entries of $Q$.

We prove that $det(Q)$ does not vanish by showing that $T^\star$ appears only once in the product of permutation $\langle 1, \ldots, n \rangle$, and that $T^\star$ does not appear in the product of any other permutation.

Before proving those facts, note that, from the conditions of lemma 2, for $1 \le i < j \le n$, paths $p_i$ and $p_j$ have no edge in common. Thus, every factor of $T^\star$ is distinct from each other.

**Proposition:** Term $T^\star$ appears only once in the product of permutation $\langle 1, \ldots, n \rangle$.

**Proof:** Let $\tau$ be a term in the product of permutation $\langle 1, \ldots, n \rangle$. Then, $\tau$ has one factor corresponding to each diagonal entry of $Q$.

A diagonal entry $q_{ii}$ of $Q$ can be expressed as a sum of three terms (see Eq.(11)).

Let $i$ be such that for all $l > i$, the factor of $\tau$ corresponding to entry $q_{ll}$ comes from the first term of $q_{ll}$ (i.e., $T(p_l)[1 + \hat{b}_{l_0}]$).

Assume that the factor of $\tau$ corresponding to entry $q_{ii}$ comes from the second term of $q_{ii}$ (i.e., $\hat{a}_{i_0 i} \cdot b_{i_0}$). Recall that each term in $\hat{a}_{i_0 i}$ corresponds to an unblocked path between $Z_i$ and $Y$, different from $p_i$, including edge $X_i \rightarrow Y$. However, from lemma 3, any such path must include either an edge which does not belong to any of $p_1, \ldots, p_n$, or an edge which appears in some of $p_{i+1}, \ldots, p_n$. In the first case, it is easy to see that $\tau$ must have a factor which does not appear in $T^\star$. In the second, the parameter of an edge of some $p_l$, $l > i$, must appear twice as a factor of $\tau$, while it appears only once in $T^\star$. Hence, $\tau$ and $T^\star$ are distinct terms.

Now, assume that the factor of $\tau$ corresponding to entry $q_{ii}$ comes from the third term of $q_{ii}$ (i.e., $\sum_{j=1}^{k} b_{i_j} \cdot a_{i_j i}$). Recall that $b_{i_j}$ is the coefficient of $\rho_{W_{i_j} Y}$ in the expression for $\phi_i(Z_i, Y, \mathbf{W}_i)$. From property $(iii)$ of lemma 1, $b_{i_j}$ is a linear function on the correlations $\rho_{Z_i W_{i_1}}, \ldots, \rho_{Z_i W_{i_k}}$, with no constant term. Moreover, correlation $\rho_{Z_i W_{i_j}}$ can be expressed as a sum of terms corresponding to unblocked paths between $Z_i$ and $W_{i_j}$. Thus, every term in $b_{i_j}$ has the term of an unblocked path between $Z_i$ and some $W_{i_j}$ as a factor. By lemma 4, we get that any such path must include either an edge that does not belong to any of $p_1, \ldots, p_n$, or an edge which appears in some of $p_{i+1}, \ldots, p_n$. As above, in both cases $\tau$ and $T^*$ must be distinct terms.

After eliminating all those terms from consideration, the remaining terms in the product of $\langle 1, \ldots, n \rangle$ are given by the expression:

$$T^* \cdot \prod_{i=1}^{n} (1 + \hat{b}_{io})$$

Since $\hat{b}_{io}$ is a polynomial on the correlations among variables $W_{i_1}, \ldots, W_{i_k}$, with no constant term, it follows that $T^*$ appears only once in this expression. $\square$

**Proposition:** Term $T^*$ does not appear in the product of any permutation other than $\langle 1, \ldots, n \rangle$.

**Proof:** Let $\pi$ be a permutation different from $\langle 1, \ldots, n \rangle$, and let $\tau$ be a term in the product of $\pi$.

Let $i$ be such that, for all $l > i$, $\pi$ selects the diagonal entry in the row $l$ of $Q$. As before, for $l > i$, if the factor of $\tau$ corresponding to entry $q_{ll}$ does not come from the first term of $q_{ll}$ (i.e., $T(p_l)[1 + \hat{b}_{l0}]$), then $\tau$ must be different from $T^*$. So, we assume that this is the case.

Assume that $\pi$ does not select the diagonal entry $q_{ii}$ of $Q$. Then, $\pi$ must select some entry $q_{il}$, with $l < i$. Entry $q_{il}$ can be written as:

$$q_{il} = b_{i0} a_{i0l} + \sum_{j=1}^{k_i} b_{i_j} a_{i_j l}$$

Assume that the factor of $\tau$ corresponding to entry $q_{il}$ comes from term $b_{i0} \cdot a_{i0l}$. Recall that each term in $a_{i0l}$ corresponds to an unblocked path between $Z_i$ and $Y$ including edge $X_l \rightarrow Y$. Thus, in this case, lemma 5 implies that $\tau$ and $T^*$ are distinct terms.

Now, assume that the factor of $\tau$ corresponding to entry $q_{il}$ comes from term $\sum_{j=1}^{k} b_{i_j} a_{i_j l}$. Then, by the same argument as in the previous proof, terms $\tau$ and $T^*$ are distinct. $\square$

Hence, term $T^*$ is not cancelled out and the lemma holds. $\square$

## 7.4 Identification of $\lambda_1, \ldots, \lambda_n$

Lemma 8 gives that $det(Q)$ is a non-trivial polynomial on the parameters of the model. Thus, $det(Q)$ only vanishes on the roots of this polynomial. However, [8] has shown that the set of roots of a polynomial has Lebesgue measure zero. Thus, system $\Phi$ has unique solution almost everywhere.

It just remains to show that we can estimate the entries of the matrix of coefficients of system $\Phi$ from data.

Let us examine again an entry $q_{il}$ of matrix $Q$:

$$q_{il} = \sum_{j=0}^{k} b_{i_j} \cdot a_{i_j l}$$

From condition $(iii)$ of lemma 1, the factors $b_{i_j}$ in the expression above are polynomials on the correlations among the variables $Z_i, W_{i_1}, \ldots, W_{i_k}$, and thus can be estimated from data.

Now, recall that $a_{i_0 l}$ is given by the sum of terms corresponding to each unblocked path between $Z_i$ and $Y$ including edge $X_l \rightarrow Y$. Precisely, for each term $t$ in $a_{i_0 l}$, there is an unblocked path $p$ between $Z_i$ and $Y$ including edge $X_l \rightarrow Y$, such that $t$ is the product of the parameters of the edges along $p$, except for $\lambda_l$.

However, notice that for each unblocked path between $Z_i$ and $Y$ including edge $X_l \rightarrow Y$, we can obtain an unblocked path between $Z_i$ and $X_l$, by removing edge $X_l \rightarrow Y$. On the other hand, for each unblocked path between $Z_i$ and $X_l$ we can obtain an unblocked path between $Z_i$ and $Y$, by extending it with edge $X_l \rightarrow Y$.

Thus, factor $a_{i_0 l}$ is nothing else but $\rho_{Z_i X_l}$. It is easy to see that the same argument holds for $a_{i_j l}$ with $j > 0$. Thus, $a_{i_j l} = \rho_{W_{i_j} X_l}$, $j = 0, \ldots, k$.

Hence, each entry of matrix $Q$ can be estimated from data, and we can solve the system of equations $\Phi$ to obtain the parameters $\lambda_1, \ldots, \lambda_n$.

## 8   Conclusion

In this paper, we presented a generalization of the method of Instrumental Variables. The main advantage of our method over traditional IV approaches, is that it is less sensitive to the set of conditional independences implied by the model. The method, however, does not solve the Identification problem. But, it illustrates a new approach to the problem which seems promising.

## Appendix

**Proof of Lemma 1:** Functions $\phi(1, \ldots, n)$ and $\psi(i_1, \ldots, i_{n-1})$ are defined recursively. For $n = 3$,

$$\begin{cases} \phi^3(1, 2, 3) & = \rho_{12} - \rho_{13}\rho_{23} \\ \psi^2(i_1, i_2) & = \sqrt{(1 - \rho_{i_1, i_2}^2)} \end{cases}$$

For $n > 3$, we have

$$
\left\{
\begin{aligned}
\phi^n(1,\ldots,n) &= \left(\psi^{n-2}(n,3,\ldots,n-1)\right)^4 \\
&\quad \cdot \phi^{n-1}(1,2,3,\ldots,n-1) \\
&\quad - (\psi^{n-2}(n,3,\ldots,n-1))^2 \\
&\quad \cdot \phi^{n-1}(1,n,3,\ldots,n-1) \\
&\quad \cdot \phi^{n-1}(2,n,3,\ldots,n-1) \\
\psi^{n-1}(i_1,\ldots,i_{n-1}) &= \left[\left(\psi^{n-2}(i_1,i_2,\ldots,i_{n-2})\right.\right. \\
&\quad \left.\cdot \psi^{n-2}(i_{n-1},i_2,\ldots,i_{n-2})\right)^2 \\
&\quad \left. - \left(\phi^{n-1}(i_1,i_{n-1},i_2,\ldots,i_{n-2})\right)^2\right]^{\frac{1}{2}}
\end{aligned}
\right.
$$

Using induction and the recursive definition of $\rho_{12.3\ldots n}$, it is easy to check that:

$$
\rho_{12.3\ldots N} = \frac{\phi^N(1,2,\ldots,N)}{\psi^{N-1}(1,N,3,\ldots,N-1)\cdot\psi^{N-2}(N,3,\ldots,N-1)}
$$

Now, we prove that functions $\phi^n$ and $\psi^{n-1}$ as defined satisfy the properties $(i) - (iv)$. This is clearly the case for $n = 3$. Now, assume that the properties are satisfied for all $n < N$.

Property $(i)$ follows from the definition of $\phi^N(1,\ldots,N)$ and the assumption that it holds for $\phi^{N-1}(1,\ldots,N-1)$.

Now, $\phi^{N-1}(1,\ldots,N-1)$ is linear on the correlations $\rho_{12},\ldots,\rho_{N-1,2}$. Since $\phi^{N-1}(2,N,3,\ldots,N-1)$ is equal to $\phi^{N-1}(N,2,3,\ldots,N-1)$, it is linear on the correlations $\rho_{32},\ldots,\rho_{N,2}$. Thus, $\phi^N(1,\ldots,N)$ is linear on $\rho_{12},\rho_{32},\ldots,\rho_{N,2}$, with no constant term, and property $(ii)$ holds.

Terms $\left(\psi^{N-2}(N,3,\ldots,N-1)\right)^2$ and $\phi^{N-1}(1,N,3,\ldots,N-1)$ are polynomials on the correlations among the variables $1,3,\ldots,N$. Thus, the first part of property $(iii)$ holds. For the second part, note that correlation $\rho_{12}$ only appears in the first term of $\phi^N(1,\ldots,N)$, and by the inductive hypothesis $\left(\psi^{N-2}(N,3,\ldots,N-1)\right)^4$ has constant term equal to 1. Also, since $\phi^N(1,2,3,\ldots,N) = \phi^N(2,1,3,\ldots,N)$ and the later one is linear on the correlations $\rho_{12},\rho_{13},\ldots,\rho_{1N}$, we must have that the coefficients of $\phi^N(1,2,\ldots,N)$ must be linear on these correlations. Hence, property $(iv)$ holds.

Finally, for property $(iv)$, we note that by the inductive hypothesis, the first term of $\left(\psi^{N-2}(N,3,\ldots,N-1)\right)^2$ has constant term equal to 1, and the second term has no constant term. Thus, property $(iv)$ holds. $\quad\square$